\documentclass[10pt,conference]{IEEEtran}
\IEEEoverridecommandlockouts
\usepackage{cite}
\usepackage{amsmath,amssymb,amsfonts}
\usepackage{graphicx}
\usepackage{textcomp}
\usepackage{algorithm}
\usepackage[noend]{algpseudocode}
\usepackage{xcolor}
\usepackage[switch]{lineno} 
\usepackage[hidelinks]{hyperref}
\usepackage{xcolor,soul,lipsum}
\newcommand{\myul}[2][black]{\setulcolor{#1}\ul{#2}\setulcolor{black}}
\newcommand{\algorithmicbreak}{\textbf{break}}
\newcommand{\BREAK}{\State \algorithmicbreak}

\newcommand\githubref[1]{%
  \renewcommand\thefootnote{}\footnote{#1}%
  \addtocounter{footnote}{-1}%
}
\usepackage{float}
\usepackage{multirow}  
\usepackage{caption}
\usepackage{subcaption}
  
\DeclareRobustCommand*{\IEEEauthorrefmark}[1]{%
\raisebox{0pt}[0pt][0pt]{\textsuperscript{\footnotesize #1}}%
}
\newcommand{\WRP}{\par\qquad\(\hookrightarrow\)\enspace}

\def\BibTeX{{\rm B\kern-.05em{\sc i\kern-.025em b}\kern-.08em
    T\kern-.1667em\lower.7ex\hbox{E}\kern-.125emX}}
\begin{document}

\title{Evaluating Online and Offline Accuracy Traversal Algorithms for k-Complete Neural Network Architectures}

\author{\IEEEauthorblockN{Yigit Alparslan\IEEEauthorrefmark{1}, Ethan Jacob Moyer\IEEEauthorrefmark{2}, Edward Kim\IEEEauthorrefmark{1}}

\IEEEauthorblockA{
\IEEEauthorrefmark{1}College of Computing \& Informatics, Drexel University, Philadelphia, PA\\
\IEEEauthorrefmark{2}School of Biomedical Engineering, Drexel University, Philadelphia, PA\\
Email: \{ ya332, ejm374, ek826 \}@drexel.edu}}

\maketitle

\begin{abstract}
Architecture sizes for neural networks have been studied widely and several search methods have been offered to find the best architecture size in the shortest amount of time possible. In this paper, we study compact neural network architectures for binary classification and investigate improvements in speed and accuracy when favoring overcomplete architecture candidates that have a very high-dimensional representation of the input. We hypothesize that an overcomplete model architecture that creates a relatively high-dimensional representation of the input will be not only be more accurate but would also be easier and faster to find. In an $N\times M$ search space, we propose an online traversal algorithm that finds the best architecture candidate in $O(1)$ time for best case and $O(N)$ amortized time for average case for any compact binary classification problem by using k-completeness as heuristics in our search. The two other offline search algorithms we implement are brute force traversal and diagonal traversal, which both find the best architecture candidate in $O(N\times M)$ time. We compare our new algorithm to brute force and diagonal searching as a baseline and report search time improvement of 52.1\% over brute force and of 15.4\% over diagonal search to find the most accurate neural network architecture when given the same dataset. In all cases discussed in the paper, our online traversal algorithm can find an accurate, if not better, architecture in significantly shorter amount of time. 
\end{abstract}


\githubref{All source code is open-sourced at \href{https://github.com/drexelai/kcompleteness-in-binary-neural-nets}{\color{blue} \myul[blue]{GitHub.}}}

\section{Introduction}

\IEEEPARstart{M}{ost} accurate neural architectures that we see today are handpicked and carefully designed by experts \cite{vgg16} \cite{resnet} to achieve the desired performance. Typically, these experts design network architectures in such a way that each hyper-parameter selection carefully supplements the problem at hand. Due a the steep learning curve for hyper-parameter selection, most novices in neural network architecture design perform a grid or brute search of possible architectures in order to locate the best performing one. This, for obvious reasons, can be slow and tedious.

However, due to the work of \cite{zophle2017} and \cite{baker2016} in the recent years, neural architecture search has seen a surge and many new traversal and search methods have been proposed. \cite{ppotuner}. The Neural Architecture Search field has seen three focus points where researchers spent most of their time. These focus points are characterized by Elsken et al, 2019 in their literature survey \cite{elsken2019} to be the following:
\begin{enumerate}
    \item \textbf{Search Space}: All potential neural architecture candidates create a search space. Depending on the problem, the search space can be quite large, hence the importance of the efficiency of the search algorithm. 
    \item \textbf{Search Algorithm}: Even though search space is usually quite large, many candidates are quite similar and, therefore, can be skipped, hence fewer candidates to look at and faster run times.
    \item \textbf{Evaluation Strategy}: Usually accuracy is given the most attention as an evaluation strategy, but depending on the problem at hand, memory consumption, individual training duration, and/or energy consumption may be a more important evaluation metric. 
\end{enumerate}{}

In this paper, we look at binary classification problems where the output layer has only one node, relating to an output classification of zero or one. We vary the number of hidden layers and the number of nodes in each layer. Such variation creates a two degrees of freedom and a N$\times$ M search space, where N is the maximum number of nodes in each layer and M is the number of hidden layers. We propose three algorithms, two of which are offline traversal algorithms and one is an online traversal algorithms.

Online traversal algorithms are those that can process input from a search space one and a time, and offline traversal algorithms are those that require all (or nearly all) of the search space to be known prior to traversing it. It becomes clear then that online algorithms are far superior than offline algorithms because they do not require the entire search space to be evaluated prior to locating the optimum. The most basic offline traversal algorithm is the brute force method where every single model candidate is evaluated and the optimum is located by comparing results over all candidates.

We mainly use the offline algorithms as a baseline for our online algorithm because these most frequently are guaranteed to locate the most optimal neural network architecture size. 

\section{Related Work}
In 2021, Alparslan et al. \cite{binarysearchinneuralnets} looked at using binary search to determine the architecture size that would give the best accuracy. There was nearly a 100-fold improvement over the naive search approach by using a modified binary search algorithm to determine the model with the highest accuracy. However, the assumptions made were too strong and worked only on certain datasets which met their criteria such as monotonic increase from both sides to global maximum for all the accuracies observed. Their work also investigated the binary search on very compact neural networks when there was only one hidden layer.  In this paper, we relax two of their assumptions: 
\begin{enumerate}
    \item Monotonic Increase from Both Sides Constraint: In our paper, we no longer require that search space would have one global maximum and values would have to increase from both sides until the global maximum is reached. \cite{binarysearchinneuralnets} 
    assumed that the search space is sorted in ascending order from beginning to the global maximum and sorted in descending order from global maximum to end. Assuming a partial sortedness on the parameter space helped apply binary search and achieve massive speed improvements in their search. However, when the underlying dataset was highly non-convex in the accuracy with respect to the architecture search space, the solution given by the search method returned a local optimum. In this paper, we no longer require such assumption.
    \item One Hidden Layer Constraint: In our paper, we are no longer constraint to just one hidden layer. We vary the number of hidden layer as well as the number of nodes in each layer. Such variation creates two degrees of freedom and helps us test our search in a much larger search space (2D).
\end{enumerate}

\section{Methodology}

\subsection{Terminology and Definitions}

In order to traverse networks with different levels of completeness, we need to have a well-defined search space and a way of representing an architecture candidate as a node in that search space. In this paper, we define a two-dimensional (2D) search space with one dimension that measures width and another that measures depth. This can be achieved by defining two terms: Initial Hidden Layer Size IHLS (\autoref{ihls}) and  Division Factor (\autoref{divisionfactor}). While the former initializes the width of the first hidden layer, the latter determines the number of subsequent hidden layers. For example, if the initial hidden layer size IHLS is 24 and the division factor DF is 3, the architecture has [24,8,2] nodes across three layers.
Defining such two terms helps representing each architecture candidate easily and well-defines the search space. 

Completeness, in short, is a measure of the depth and width of a given network architecture. If the hidden layer dimensions are greater than that of the input layer, we say the model incurs an overcomplete representation, and if it is less it is an undercomplete representation \cite{overcompleteness}. In the search space, some architecture candidates are overcomplete and some are undercomplete. In order to distinguish the variation among them, we define a k-completeness score in \autoref{completenessScore}. The intuition behind it is to assign the size of the next hidden layer with quotient of a given hidden layer size and the value of the division factor. This score aims to distinguish overcomplete architectures from undercomplete architectures and to illustrate the trend between k-completeness and training time for accurate models.

In order to formally define a k-completeness score that ranges from zero to infinity, we provide the following generalized definitions for jumping factor (\autoref{jumpingfactor}) and the division factor (\autoref{divisionfactor}). 
\newline\newline
\textbf{Initial Hidden Layer Size:}
\begin{equation}\label{ihls}
\centering
\begin{aligned}
    IHLS = size(Hidden Layer _{1})
\end{aligned}
\end{equation}

\textbf{Division Factor:}
\begin{equation}\label{divisionfactor}
\centering
\begin{aligned}
    Division Factor =\frac{size(Layer(i))}{size(Layer(i+1))}, \\ 1\le i\le N-1
\end{aligned}
\end{equation}

\textbf{Jumping Factor:}
\begin{equation}\label{jumpingfactor}
\centering
\begin{aligned}
    Jumping Factor = \frac{IHLS}{size(Input Layer)}
\end{aligned}
\end{equation}

\textbf{k-Completeness Score:}
\begin{equation}\label{completenessScore}
\centering
\begin{aligned}
    k\textendash Completeness = \alpha \times jumping\_factor + \\
    (1-\alpha)\times\frac{1}{DF}
\end{aligned}
\end{equation}

The $\alpha$ in \autoref{completenessScore} for our experiment is chosen to be 0.5 because we wanted to weight the division factor and the jumping factor equally.

\begin{figure}[H]
\centerline{\includegraphics[width=\linewidth]{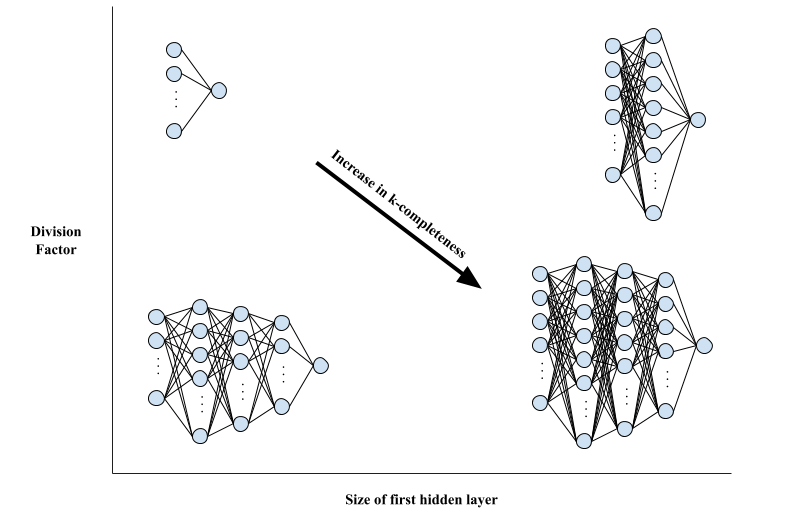}}
\caption{Search space for neural network architecture k-completeness as it varies with the size of the first hidden layer (x-axis) and the division factor (y-axis). 
On top left corner, a very large division factor DF (see \autoref{divisionfactor}) and a very small initial hidden layer size IHLS (see \autoref{ihls}) are resulting in very undercomplete models. On bottom right, a very small division factor and a very large initial hidden layer size are resulting in very overcomplete models
}
\label{completenesssearchfig}
\end{figure}

The interplay between these two factors and the width and depth of a network is not apparently obvious. While the size of the first hidden layer is proportional the k-completeness of the network, the division factor is inversely proportional. In other words, a large division factor yields a small number of subsequent hidden layers as there is a larger value by which each hidden layer is divided by in order to yield the next. This space, along with classifications of low and high k-completeness score, is observed in \autoref{completenesssearchfig}.

\subsection{Traversal Algorithms}
\begin{figure}[htb]
\centerline{\includegraphics[width=0.9\linewidth]{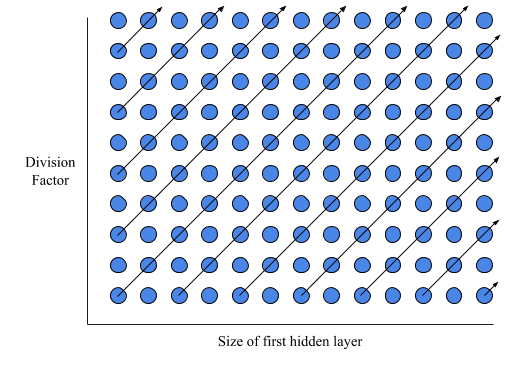}}
\caption{Diagonal traversal algorithm. This accuracy traversal algorithm skips nodes that are closer to each other in the same row and completes faster than the naive brute force approach. }
\label{diagonalfig}
\end{figure}

In Figures \ref{bruteforcefig}, \ref{diagonalfig}, \ref{zigzagfig}, the y-axis represents division factor DF and is assigned powers of 2 as values (2, 4, 8, 16, etc.). The x-axis represents the initial hidden layer size IHLS and is assigned values that range from 1 to the maximum IHLS. For example, in Figure \ref{bruteforcefig}, there are 10 rows, so the y-axis goes from 2, 4, 8, 16, ..., to 1024 from bottom left to top left. There are 10 columns so the x-axis goes from 1 to 10 from bottom left to bottom right. Such set up well-defines the search space and facilitates easier implementation since each node represents an architecture candidate. For example, the node labeled \textbf{B} has DF of two and IHLS of 10, which designates the architecture [10,5,2,1]. The node labeled \textbf{A} DF of two and IHLS of nine, which designates the architecture [9,4,2,1]. The node labeled \textbf{C} has DF of four and IHLS of ten, which designates the architecture [10,2] as defined in \autoref{ihls} and \autoref{divisionfactor}. 
\subsubsection{Offline Algorithm \#1: Brute Force Search}

\begin{figure}[htb]
\centerline{\includegraphics[width=\linewidth]{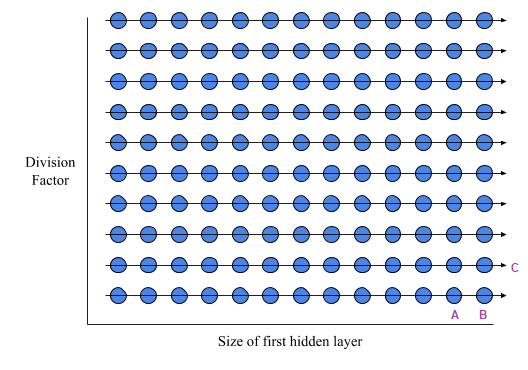}}
\caption{Brute force traversal algorithm. This accuracy traversal is considered offline because traversal does not require other architectures' positions.   }
\label{bruteforcefig}
\end{figure}

This first offline algorithm is analogous to a linear search for a maximum value in a list--each value in the search space (or list) needs to be evaluated before a result can be determined. For this reason, in the case of the architecture search space, the complexity of this algorithm in best case, worst case, and average case is $O(N \cdot M)$. For hyper-parameter selection, this complexity is terribly large because each time the model architecture is changed slightly, the model has to be completely reevaluated. Regardless, this greedy algorithm is displayed in \autoref{bruteforcefig}.

\subsubsection{Offline Algorithm \#2: Diagonal Search} The diagonal traversal algorithm returns alternating primary diagonals. In this way, this offline algorithm does not evaluate all of the network architectures. Instead, it generalizes that the difference between the global optimum and the closest optimum is negligible. Regardless, similar to the brute force search, this diagonal search has a complexity of $O(N^2)$ in the best case, worst case, and average case. 

\begin{algorithm}[t]
\caption{Returns search space marked by alternating diagonals.}
\begin{algorithmic}[1]
\Procedure{\textit{DiagonalSearch}}{$space$}
\State $\textit{D} \gets []$
\For{$i$ := 1 to $length(space)$}
\For{}{$i$ := 1 to $length(space[0])$}
\If{$i + j$ mod $2$}
\State $\textit{D.append(space[i][j])}$
\EndIf
\EndFor
\EndFor
\Return $D$
\EndProcedure
\end{algorithmic}
\end{algorithm}

\subsubsection{Online Algorithm \#1: Zigzag Search}
\begin{figure}[!ht]
\centerline{\includegraphics[width=0.8\linewidth]{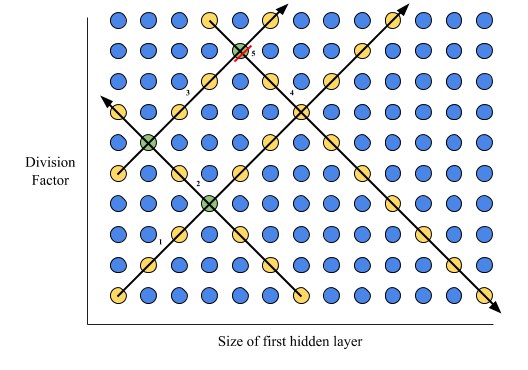}}
\caption{Zigzag traversal algorithm. This algorithm relies on traversing the search space with alternating primary and secondary diagonals. Blue circles represent unseen network architectures, yellow represents seen network architectures, and greens represent optimal network architectures along a diagonal. The first pass forms a primary diagonal stretching along an indeterminate change of completeness from the lower left and corner to the upper right hand corner of the search space. It locates an optimal architecture four nodes into the search. The second pass forms a smaller secondary diagonal line which is used to locate the second optimal architecture along that determinate diagonal. From here, a third pass is the second primary diagonal, which is much smaller than the first. The search ends once the fourth diagonal finds a optimal architecture that was already previously recorded, such as in this case on the third diagonal line.}
\label{zigzagfig}
\end{figure}

Zigzag traversal algorithm is the final algorithm that we investigate in this paper. It is an online traversal algorithm, meaning that the algorithm finds the next candidate by only processing the current candidates. By skipping over candidates similar in architecture, this algorithm sees significant running time improvements, even though it becomes harder to implement. \autoref{zigzagfig} explains the steps visually and Algorithm~\ref{alg:zigzag} formalizes the algorithm. Going in the opposite diagonal after finding the best architecture along that diagonal has its intuition coming from gradient descent. The opposite direction intuitively represents the orthogonal direction of the gradients along the surface when approaching a local minimum. The success of this accuracy traversal algorithm (see \autoref{table:results}  for results) can  be attributed to the fact that every time we go in the opposite diagonal, we are preventing oscillation \cite{oscillation} \cite{oscillation2} and vanishing gradients \cite{vanishinggradients} and providing a random jump factor to avoid getting stuck on local solutions. 

\begin{algorithm}[ht]
\caption{Performs an online traversal of the search space using alternating primary and secondary diagonals to find the optimal network architecture.}
\label{alg:zigzag}

\begin{algorithmic}[H]

\Procedure{\textit{ZigZagSearch}}{$space$}
\State $\textit{visited} \gets []$
\State $\textit{search} \gets True$
\State $\textit{isPrimary} \gets True$
\State $\textit{start} \gets (0, 0)$

\State $\textit{max\_accuracy} \gets 0$
\State $\textit{optimal\_param} \gets (0 ,0)$

\While{$search$}
\If{$\textit{isPrimary}$}
\For{$(x, y)$ in \WRP $getPrimaryDiagonal(space, start)$}
\State $\textit{current\_accuracy} \gets \textit{Pipeline(model(x, y))}$
\State $\textit{visited.append((x, y))}$
\If{$\textit{current\_accuracy} > \textit{max\_accuracy}$}
\State $\textit{max\_accuracy} \gets current\_accuracy$
\State $\textit{optimal\_param} \gets (x ,y)$
\State $\textit{start} \gets (x ,y)$
\EndIf
\EndFor

\If{$sum(\textit{optimal\_param} == visited) == 2$}
\State $\textit{search} \gets False$
\BREAK
\EndIf
\State $\textit{isPrimary} \gets False$
\Else
\For{$(x, y)$ in \WRP $getSecondaryDiagonal(space, start)$}
\State $\textit{current\_accuracy} \gets \textit{Pipeline(model(x, y))}$
\State $\textit{visited.append((x, y))}$
\If{$\textit{current\_accuracy} > \textit{max\_accuracy}$}
\State $\textit{max\_accuracy} \gets current\_accuracy$
\State $\textit{optimal\_param} \gets (x ,y)$
\State $\textit{start} \gets (x ,y)$
\EndIf
\EndFor
\If{$sum(\textit{optimal\_param} == visited) == 2$}
\State $\textit{search} \gets False$
\BREAK
\EndIf
\State $\textit{isPrimary} \gets True$
\EndIf

\EndWhile
\Return ($max\_accuracy$, $optimal\_param$)
\EndProcedure
\end{algorithmic}
\end{algorithm}

\begin{table}[H]
\caption{Comparison among all traversal algorithms. For all cases, search space is considered to be a N$\times$M matrix. Zigzag search has the same worst case scenario as other naive approaches. However, zigzag search has a cheaper amortized cost at each traversal. }
\begin{center}
\resizebox{\columnwidth}{!}{%
\begin{tabular}{|c|c|c|c|}
\hline
\textbf{Traversal}&\multicolumn{3}{|c|}{\textbf{Running Time Complexity}} \\
\cline{2-4} 
\textbf{Algorithms} & \textbf{\textit{Best Case: $\Omega$}}& \textbf{\textit{Worst Case: O}}& \textbf{\textit{Average Case:$\Theta$}} \\
\hline
Brute Force Search &  $O(N\cdot M)$ &  $O(N\cdot M)$ & $O(N\cdot M)$\\
\hline
Diagonal Search &  $O(N\cdot M)$ &  $O(N\cdot M)$ & $O(N\cdot M)$\\
\hline
Zigzag Search &  $O(1)$ &  $O(N\cdot M)$ &  $O(N) (Amortized) $\\
\hline
\end{tabular}
}
\label{tab1}
\end{center}
\end{table}

\begin{figure*}
\centerline{\includegraphics[width=\textwidth]{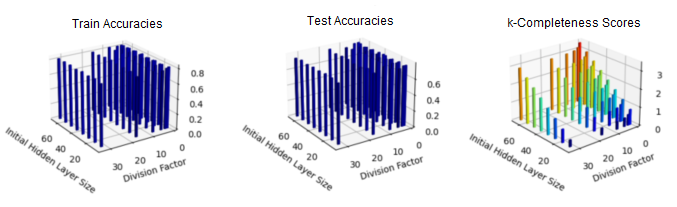}}
\caption{Titanic Model Train Accuracies, Test Accuracies and k-Completeness Scores. Train and test accuracies have very small dissimilarities for Titanic model. Very small division factor and very large initial hidden layer size results in overcomplete architectures, which have very large k-completeness scores.}
\label{churn_model_results}
\end{figure*}

\begin{figure*}
\centerline{\includegraphics[width=\textwidth]{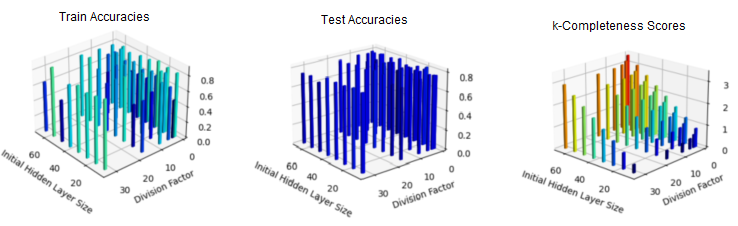}}
\caption{Churn Model Train Accuracies, Test Accuracies and k-Completeness Scores. For Churn model, over complete architectures, which have very large k-completeness scores perform better train and testing accuracies.}
\label{titanic_model_results}
\end{figure*}

\subsection{Datasets}
In this study, we use two datasets, titanic dataset and customer churn dataset. 
\subsubsection{Titanic dataset}
This dataset is made public by Kaggle\cite{titanicdataset}. The dataset has 14 columns to indicate each passenger on the Titanic ship. The columns include sex, name, destination, fare, destination etc. The dataset has about 1000 rows. The label for each row has a 1 or 0 to indicate whether the passenger survived the crash or not, hence the binary classification. The architecture candidates that we search to predict this binary classification problem has all 11 dimensions in the input layer and 1 in the output layer. 

\subsubsection{Churn dataset}
This dataset is made public by Drexel Society of Artificial Intelligence \cite{churndataset}. The dataset has information about customers that are using an imaginary contract and has labels to indicate whether the company has left the contract or not.  The model that we build for this dataset tries to predict whether the customer is about to leave this contract or not, hence the binary classification. There are 14 columns and 10000 rows in the dataset. The dataset columns has features to indicate the customer revenue, customer contract cost, the type of the product the contract is about, the region of the customer, the fact that the customer whether renewed the contract in the previous 90 days or so etc. The dataset has features represented as categories as well as floating point numbers, therefore, we have to scale and apply categorical transformation when we train a model for this dataset. All the architectures that we look in the search space has 11 dimensions in the input layer and 1 in the output layer.

\section{Results and Observations}

All three traversal algorithms have been run against two models. Zigzag traversal algorithm has been found to be the fastest to find a model architecture candidate among all the traversal algorithms. The online nature of algorithm and the low amortized cost of finding a new candidate yields significant improvements in the running time. Zigzag traversal might get stuck on a local optimum and miss the global optimum due to its online nature. We see that the training accuracy and the testing accuracy of the candidate found by zigzag traversal is on average 1.49\% and 1.52\% lower for the titanic model respectively compared to brute force and diagonal traversal. For the titanic model, the brute force traversal was able to find an architecture candidate about 7.4\% more sparse than the zigzag traversal algorithm, which was surprising since the brute force doesn't take the k-completeness into consideration, but the zigzag traversal algorithm takes the favors sparse architectures. The resulting architecture of the zigzag traversal was about 5\% more sparse than that of the diagonal traversal. 

Overall, for the \textbf{titanic model}, zigzag traversal algorithm found an architecture with a training accuracy of 80.89\% and a testing accuracy of 77.48\%, which is \textbf{only 1.49\% lower} than the average training accuracy of the other traversal algorithms and \textbf{only 1.52\% lower} than the average testing accuracy of the other traversal algorithms, but the completion time for the zigzag traversal is about \textbf{2 times faster than the brute force and 1.18 times faster than the diagonal search.} Due to relatively small size of the titanic dataset(1000 rows) compared to the churn dataset(10000 rows), running time improvements when zigzag traversal are not truly observed.

We see a slightly better performance in the training and testing accuracy for zigzag traversal when applied to the churn model. The training accuracy of the candidate found by the zigzag traversal is 3.95\% lower than that of the brute force traversal, but 3.34\% better than that of the diagonal traversal. The testing accuracy of the candidate found by the zigzag traversal is very similar to that of the brute force and diagonal traversal, being only 0.84\% and 0.69\% lower compared to them. The candidate found by the zigzag traversal for the churn model is the most sparse architecture model compared to that of brute force and diagonal traversal. In fact, the zigzag traversal was able to find a candidate with a k-completeness score twice (2.18x) as much as that of the brute force. The high k-completeness score for the result of the zigzag traversal is the result of the nature of the algorithm where each traversal in the primary diagonal is followed by a traversal in the secondary diagonal to skip over similar and dense architectures and favor the more sparse ones over less sparse ones. 

Overall, for the \textbf{churn model}, zigzag traversal algorithm found an architecture with a training accuracy of 79.98\% and a testing accuracy of 81.00\%, which is \textbf{only 0.305\% lower} than the average training accuracy of the other traversal algorithms and \textbf{only 0.765\% lower} than the average testing accuracy of the other traversal algorithms, but the completion time for the zigzag traversal is about \textbf{5 times faster than the brute force and 3 times faster than the diagonal search.} One more thing to note is that the architecture found by the churn model is not only more sparse than the result of the other traversal algorithms, but about 30\% more sparse than the second most sparse architecture that the zigzag traversal has ever discovered during its search. In this section, we only report the plots for the zigzag accuracy traversal algorithm. The reader can see ~\autoref{appendix} for the plots of other traversal algorithms.

\begin{figure}[!ht]
\centerline{\includegraphics[width=\linewidth]{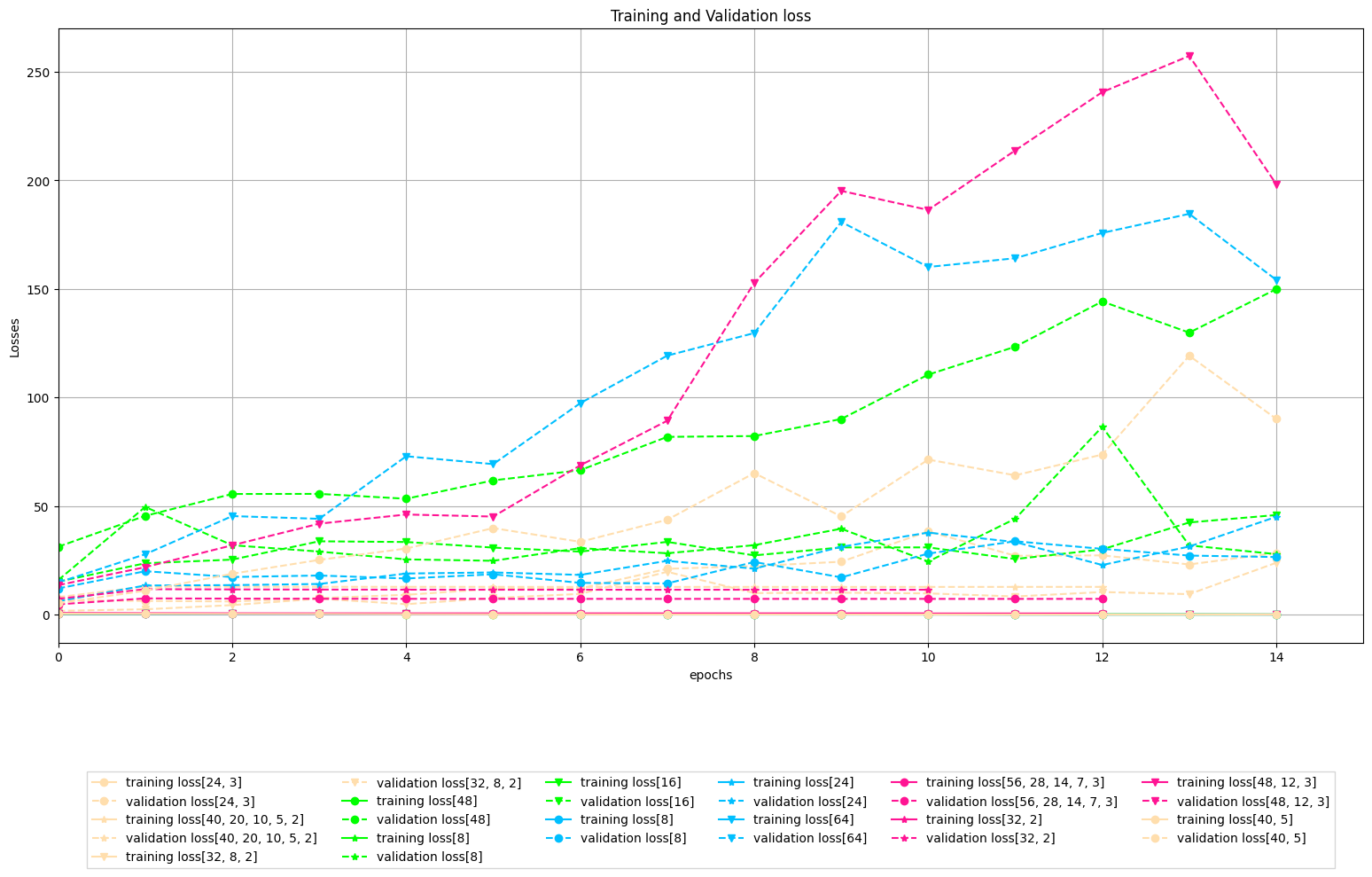}}
\caption{Loss Plots of Models found by Zigzag Traversal for Churn Model}
\label{churn_zigzag_search_loss}
\end{figure}

\begin{figure}[!ht]
\centerline{\includegraphics[width=\linewidth]{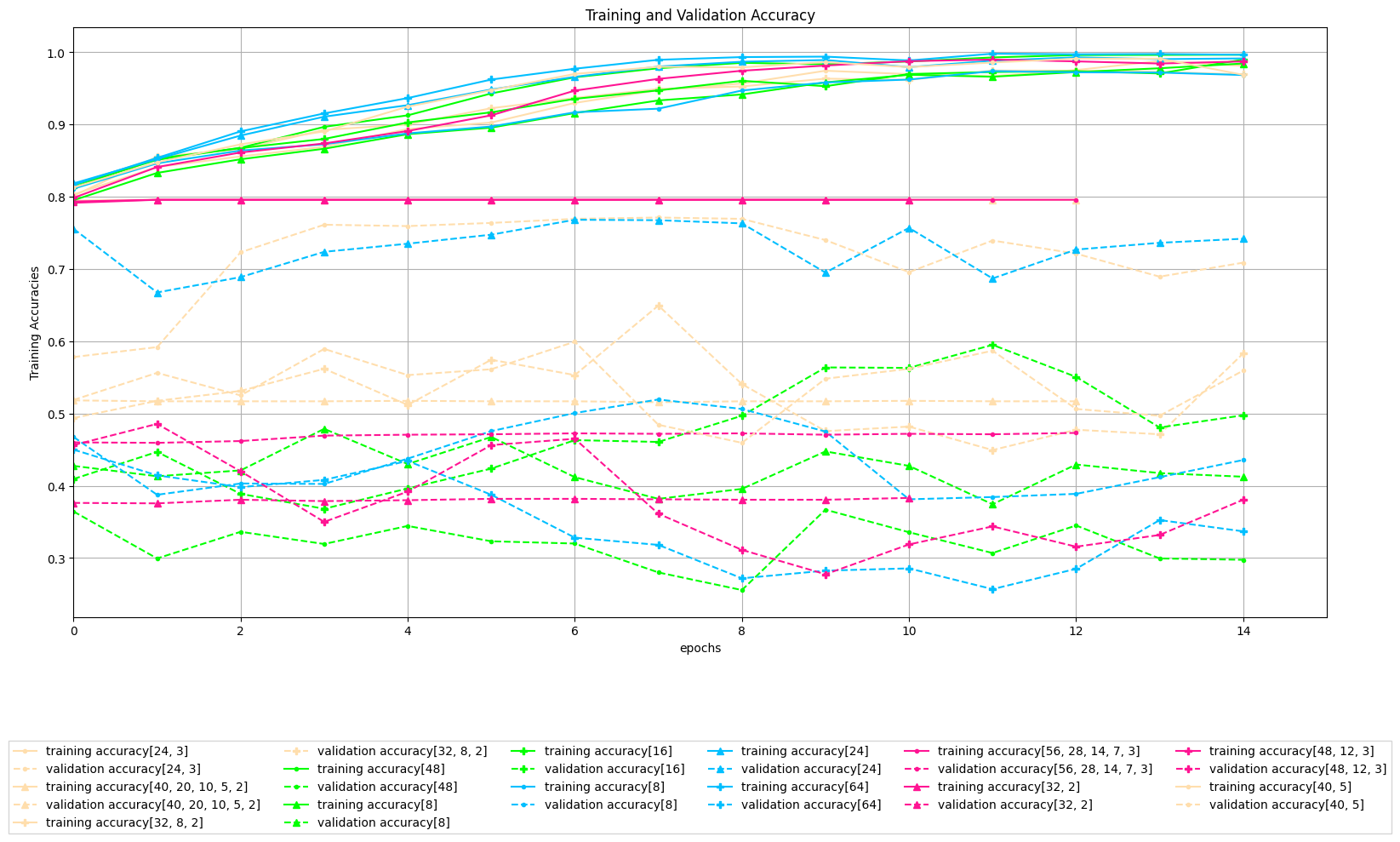}}
\caption{Accuracy Plots of Models found by Zigzag Traversal for Churn Model.  }
\label{churn_zigzag_search_accuracy}
\end{figure}

\begin{figure}[!ht]
\centerline{\includegraphics[width=\linewidth]{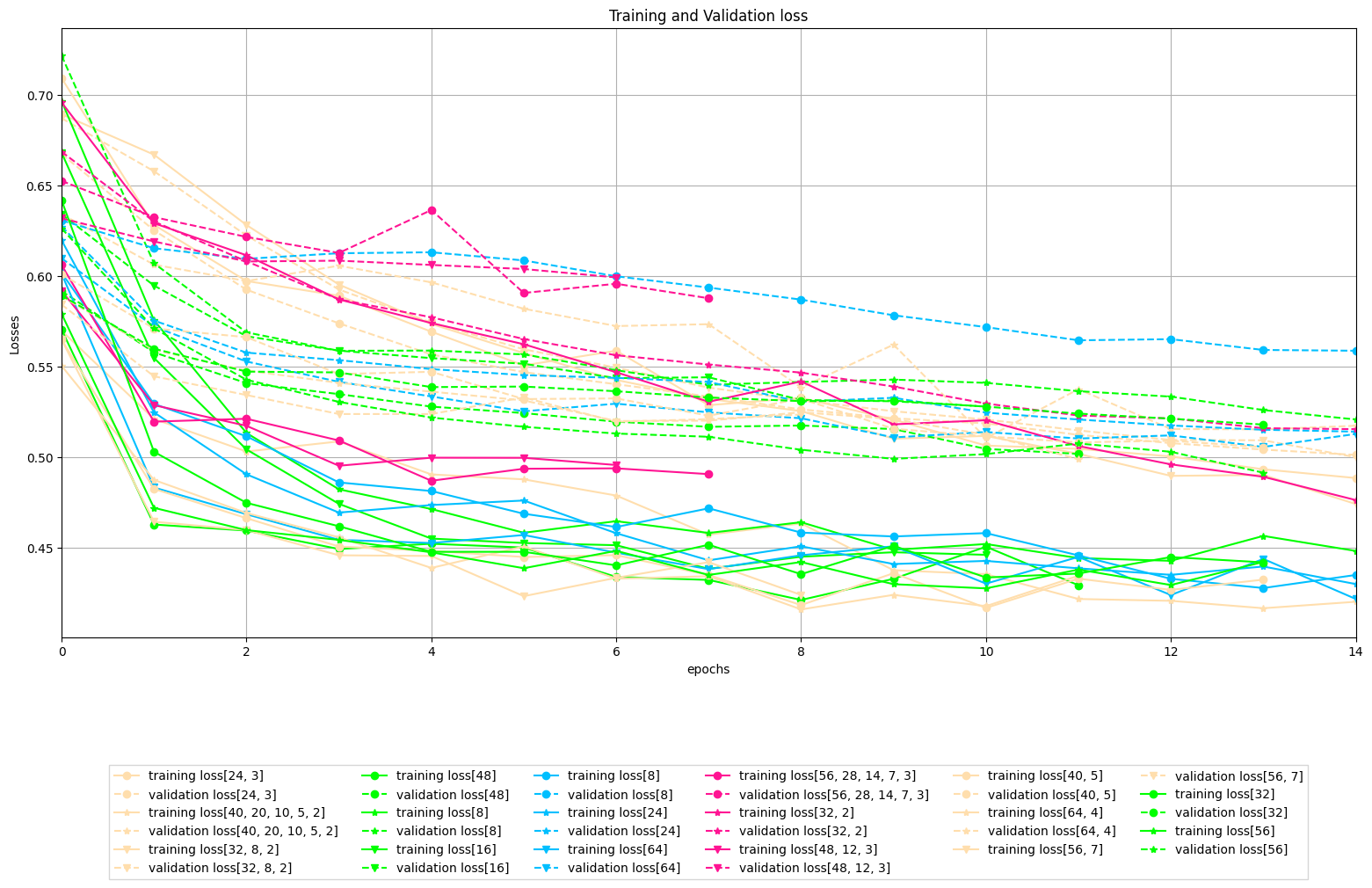}}
\caption{Loss Plots of Models found by Zigzag Traversal for Titanic Model}
\label{titanic_zigzag_search_loss}
\end{figure}

\begin{figure}[!ht]
\centerline{\includegraphics[width=\linewidth]{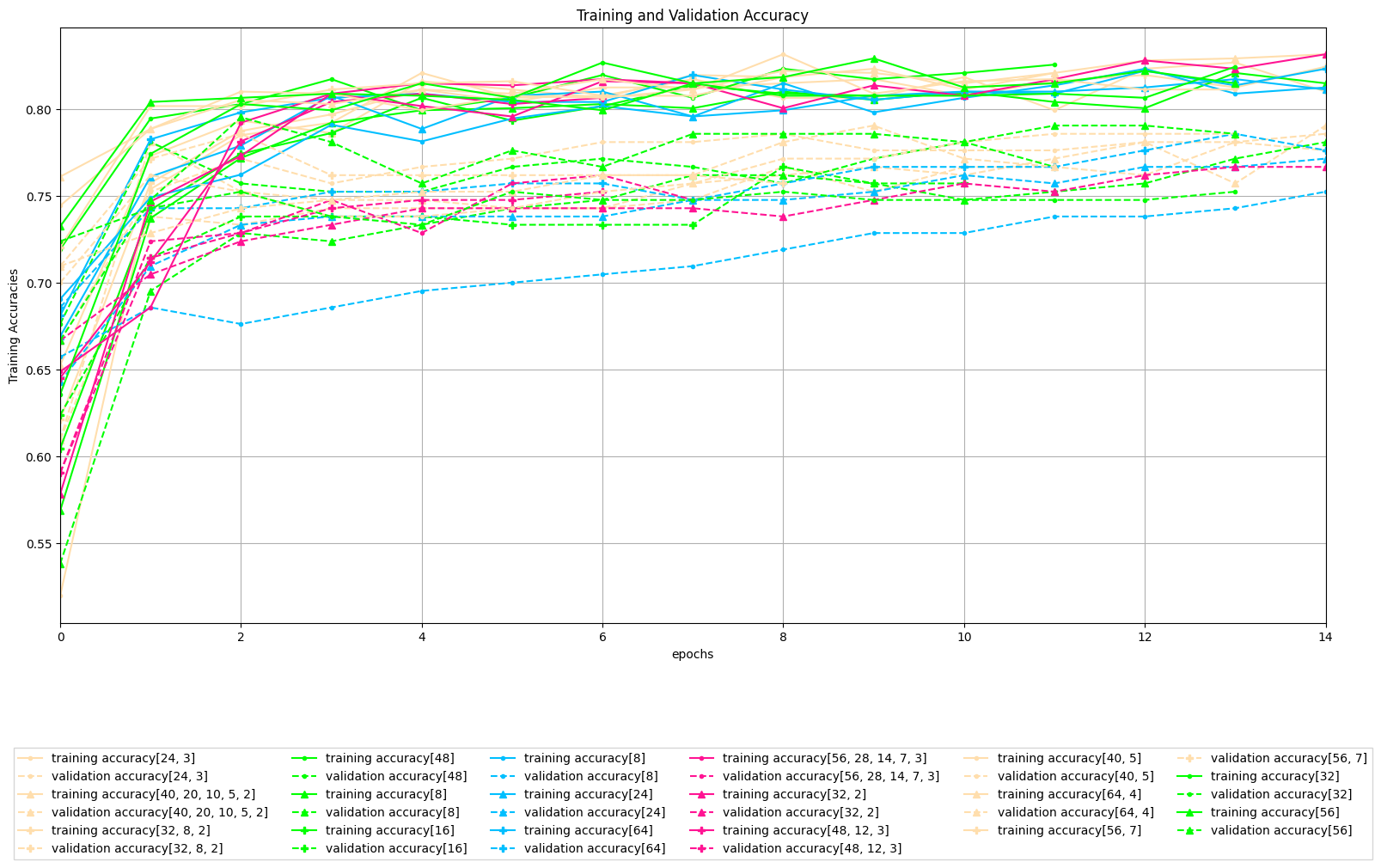}}
\caption{Accuracy Plots of Models found by Zigzag Traversal for Titanic Model}
\label{titanic_zigzag_search_accuracy}
\end{figure}

\begin{table}[!ht]
\centering
\caption{Results for All Traversal Algorithms. For all cases, initial hidden layer size is 64 and the division factor is 64. All comparison programs are run on Intel Core i5 4-cores 1.7 GHz processor with 16 GB memory and are written in Python 3.7. Completion times are reported in seconds.}
\resizebox{\columnwidth}{!}{\begin{tabular}{cc|c|c|c|}
\hline
\multicolumn{1}{|c|}{\textbf{Model}} &
  \textbf{Metrics} & \textbf{Brute Force} & \textbf{Diagonal} & \textbf{Zigzag} \\ \hline

\multicolumn{1}{|c|}{\multirow{5}{*}{\textbf{Titanic}}} &
  \textbf{Completion Time} & 139.098 & 78.695 & \textbf{66.6}\\ 
\multicolumn{1}{|c|}{} & \textbf{Train Acc.} & 82.43\% & 82.33\% & 80.89\%\\ 
\multicolumn{1}{|c|}{} & \textbf{Test Acc.} & 78.62\% & 79.38\% & 77.48\%\\ 
\multicolumn{1}{|c|}{} & \textbf{k-Completeness} & 3.1591 & 2.7955 & 2.9403\\ 
\multicolumn{1}{|c|}{} & \textbf{Best Architecture} & [64,32,16,8,4,2] & [56,28,14,7,3] & [64,4]\\
\hline

\multicolumn{1}{|c|}{\multirow{5}{*}{\textbf{Churn}}} &
  \textbf{Completion Time} & 3854.482 & 1989.97 & \textbf{672.0}\\
\multicolumn{1}{|c|}{} & \textbf{Train Acc.} & 83.93\% & 76.64\% & 79.98\%\\ 
\multicolumn{1}{|c|}{} & \textbf{Test Acc.} & 81.84\% & 81.69\% & 81.00\%\\
\multicolumn{1}{|c|}{} & \textbf{k-Completeness} & 1.2273 & 1.8807 & 	2.6818 \\
\multicolumn{1}{|c|}{} & \textbf{Best Architecture} & [16] & [40, 5] & [48] \\ \hline
\end{tabular}}
\label{table:results}
\end{table}

\section{Conclusion}
In this paper, we proposed an online traversal algorithm to find the best architecture candidate in a search space in O(1) time for best case and O(N) amortized time for average case for any compact binary classification problem by using k-completeness as heuristics in our search. We compared our new algorithm to brute force and diagonal searching as a baseline and reported search running time improvement of 52.1\% over brute force and of 15.4\% over diagonal search to find the most accurate neural network architecture when given the same dataset. Our online traversal algorithm could find accurate architectures that were on par, if not better, than the other algorithms that we discuss in this paper. We hope that our findings will give insights to researchers in the Neural Architecture Search field when performing exhaustive grid search to find the most accurate architectures in shortest amount of time possible.

\section{Future Work}

We hope to develop more online algorithms. This is because a different online algorithm might prove be more efficient with certain search spaces or architecture sizes compared to the three algorithms explored in this paper. Additionally, the models that we study in this paper are compact models with no convolutional layers. Moving from binary classification problems to image recognition and seeing the effect of our online algorithm might be an interesting study in the future. 

\section{Acknowledgment}

We would like to acknowledge Drexel Society of Artificial Intelligence for its contributions and support for this research.

\section*{Appendix}
\label{appendix}
\begin{figure}[h!]
\centerline{\includegraphics[width=0.92\linewidth]{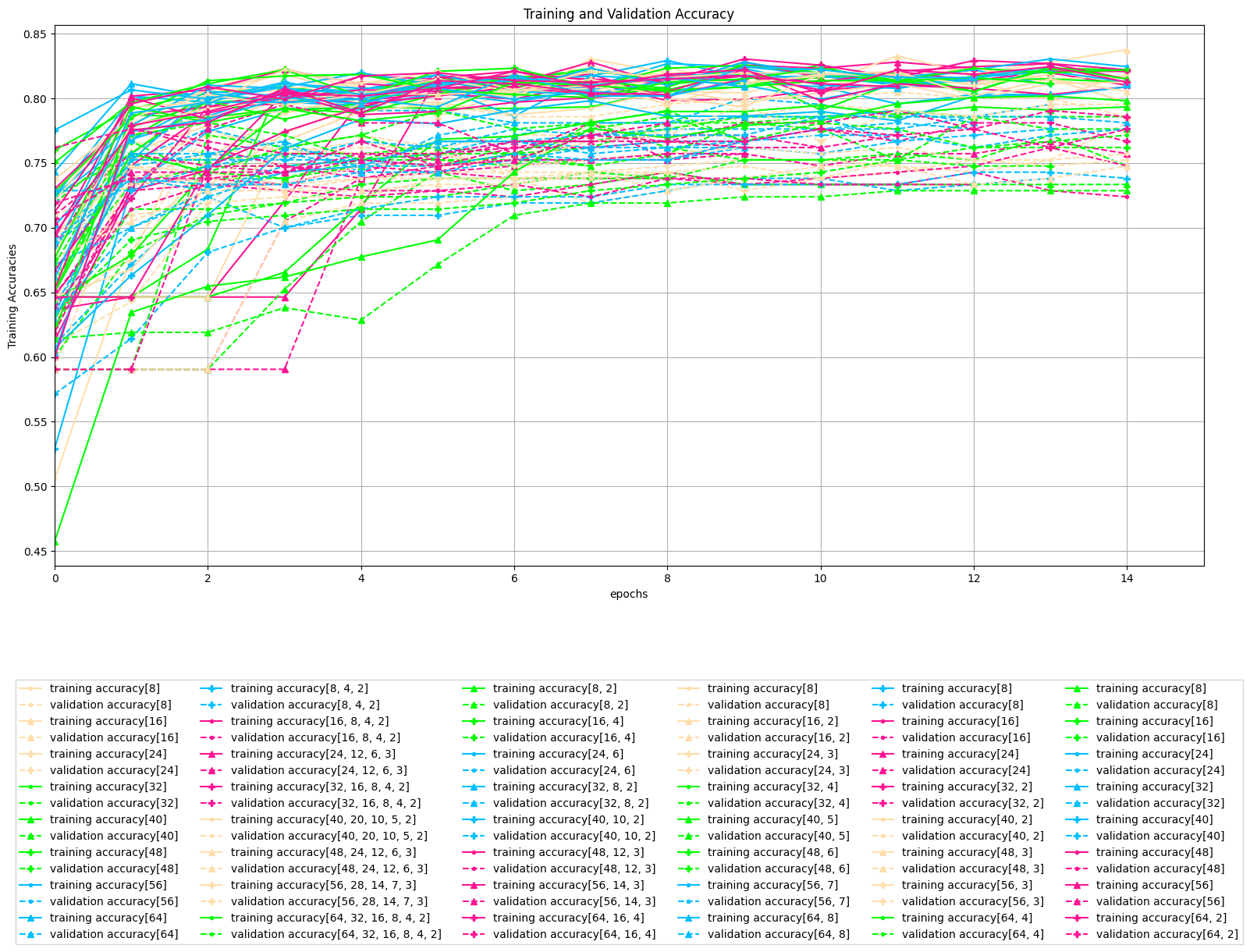}}
\caption{Accuracy Plots of Models found by Brute Force Traversal for Titanic Model}
\label{titanic_brute_force_search_accuracy}
\end{figure}

\begin{figure}[b!]
\centerline{\includegraphics[width=0.92\linewidth]{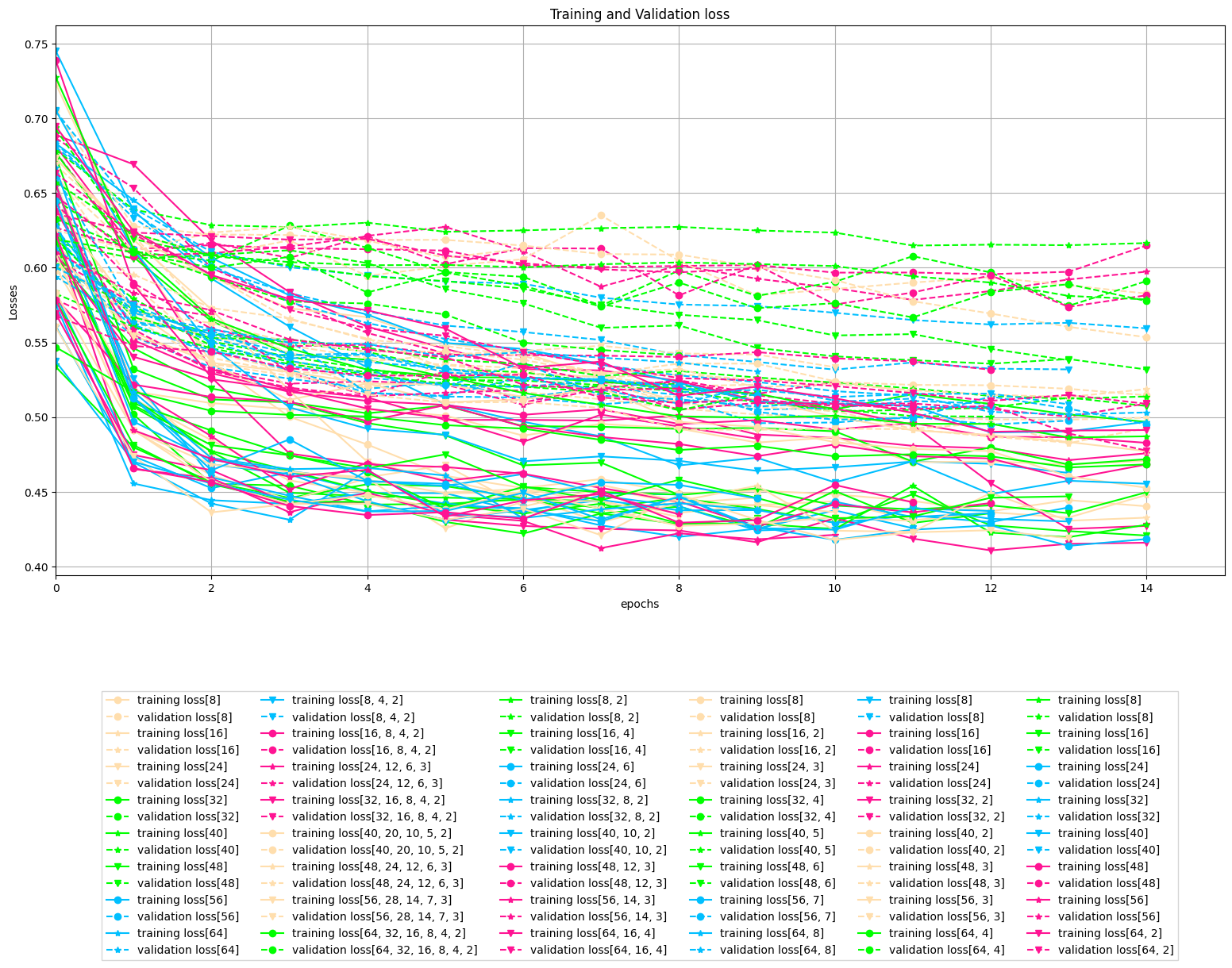}}
\caption{Loss Plots of Models found by Brute Force Traversal for Titanic Model}
\label{titanic_brute_force_search_loss}
\end{figure}

\begin{figure}[htbp!]
\centerline{\includegraphics[width=\linewidth]{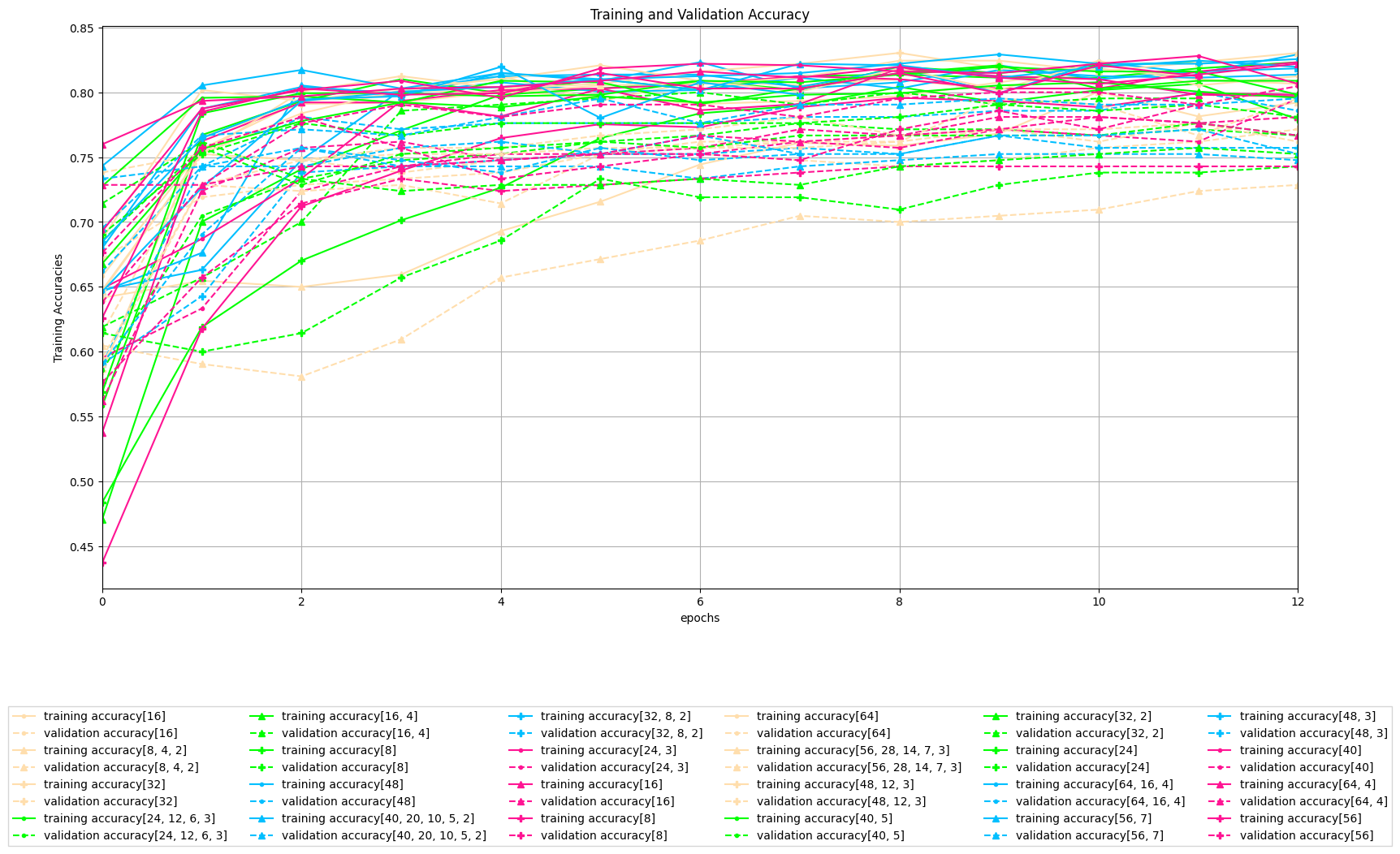}}
\caption{Accuracy Plots of Models found by Diagonal Traversal for Titanic Model}
\label{titanic_diagonal_search_accuracy}
\end{figure}

\begin{figure}[htbp!]
\centerline{\includegraphics[width=\linewidth]{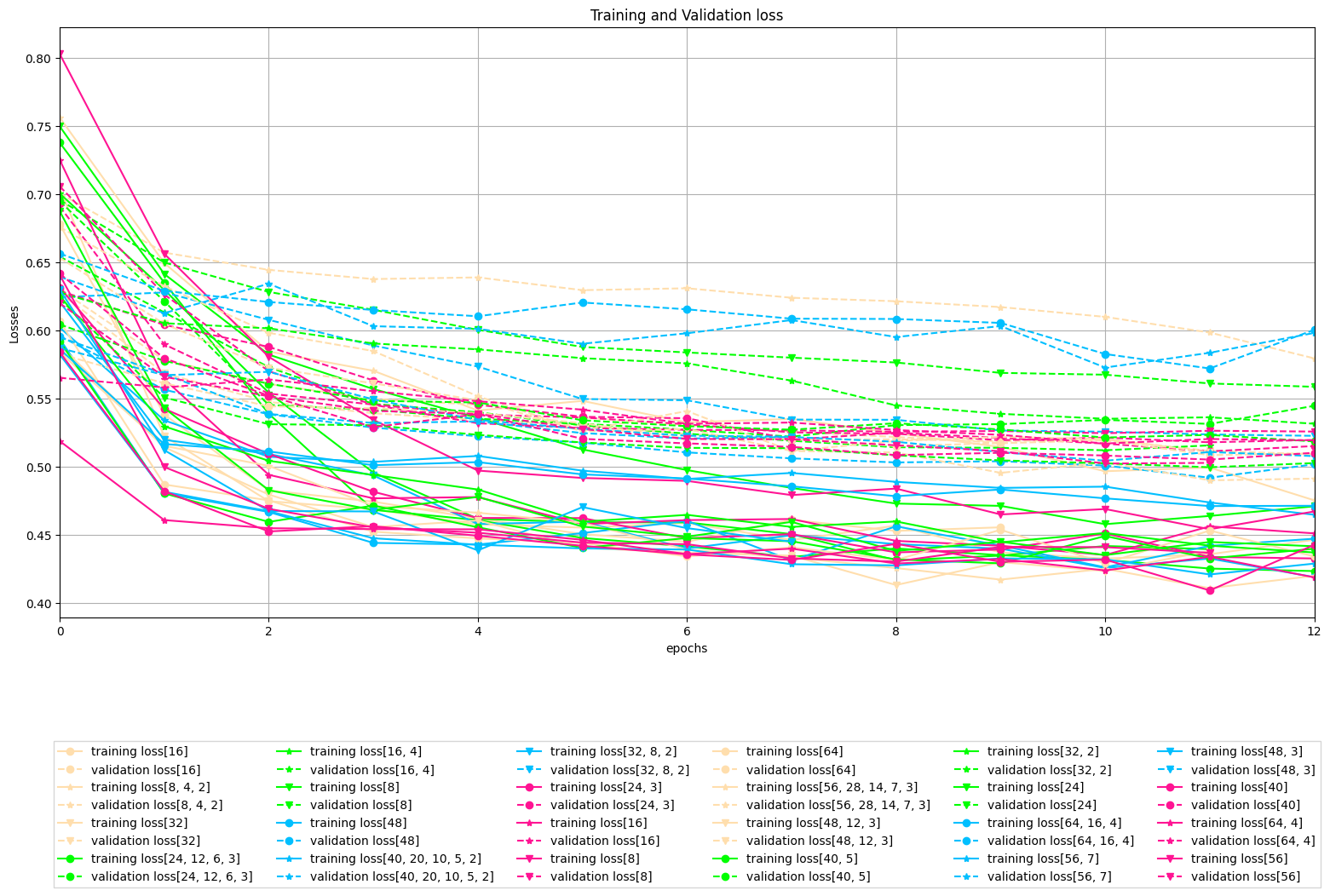}}
\caption{Loss Plots of Models found by Diagonal Traversal for Titanic Model}
\label{titanic_diagonal_search_loss}
\end{figure}

\begin{figure}[htbp!]
\centerline{\includegraphics[width=\linewidth]{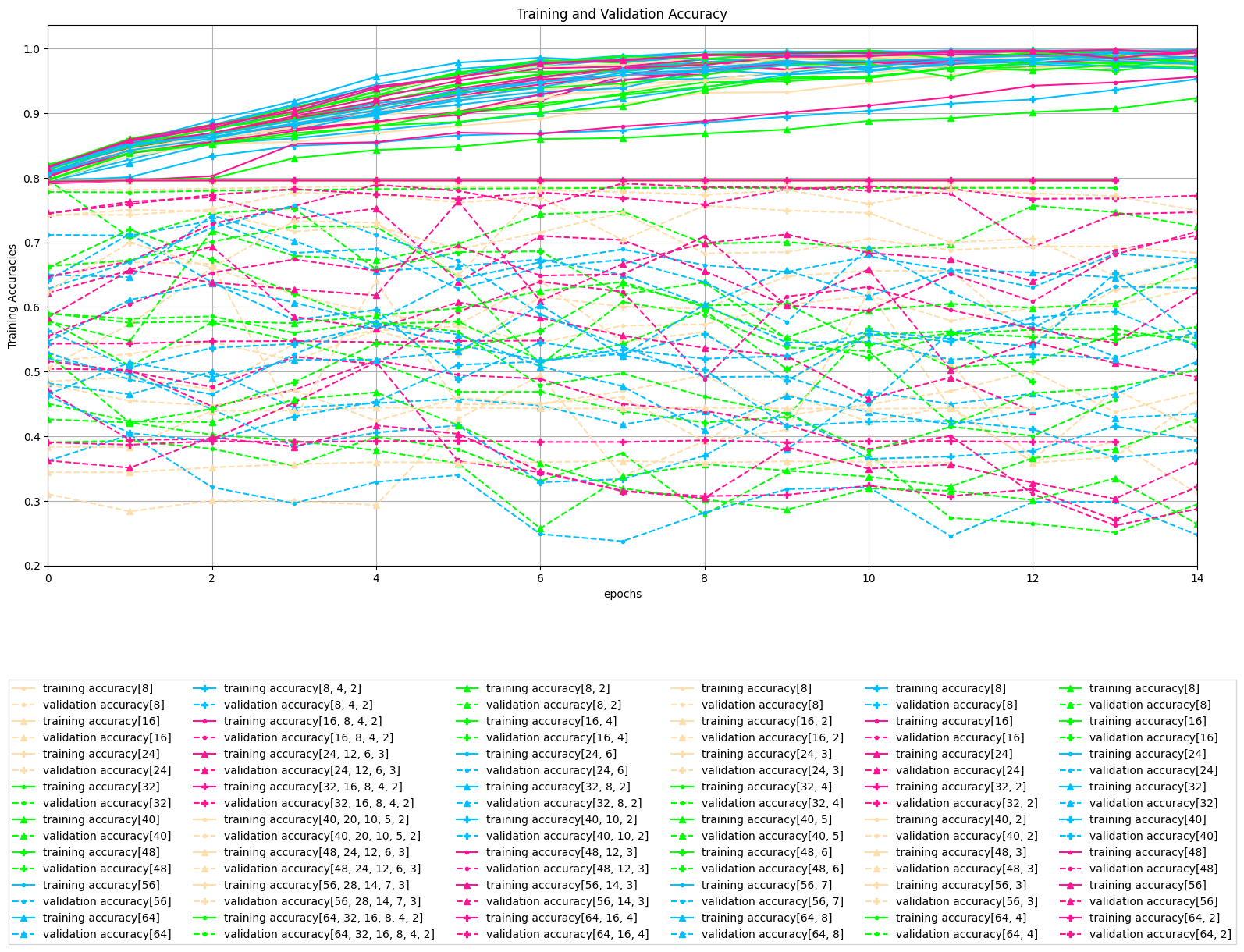}}
\caption{Accuracy Plots of Models found by Brute Force Traversal for Churn Model}
\label{churn_brute_force_search_accuracy}
\end{figure}

\begin{figure}[htbp!]
\centerline{\includegraphics[width=\linewidth]{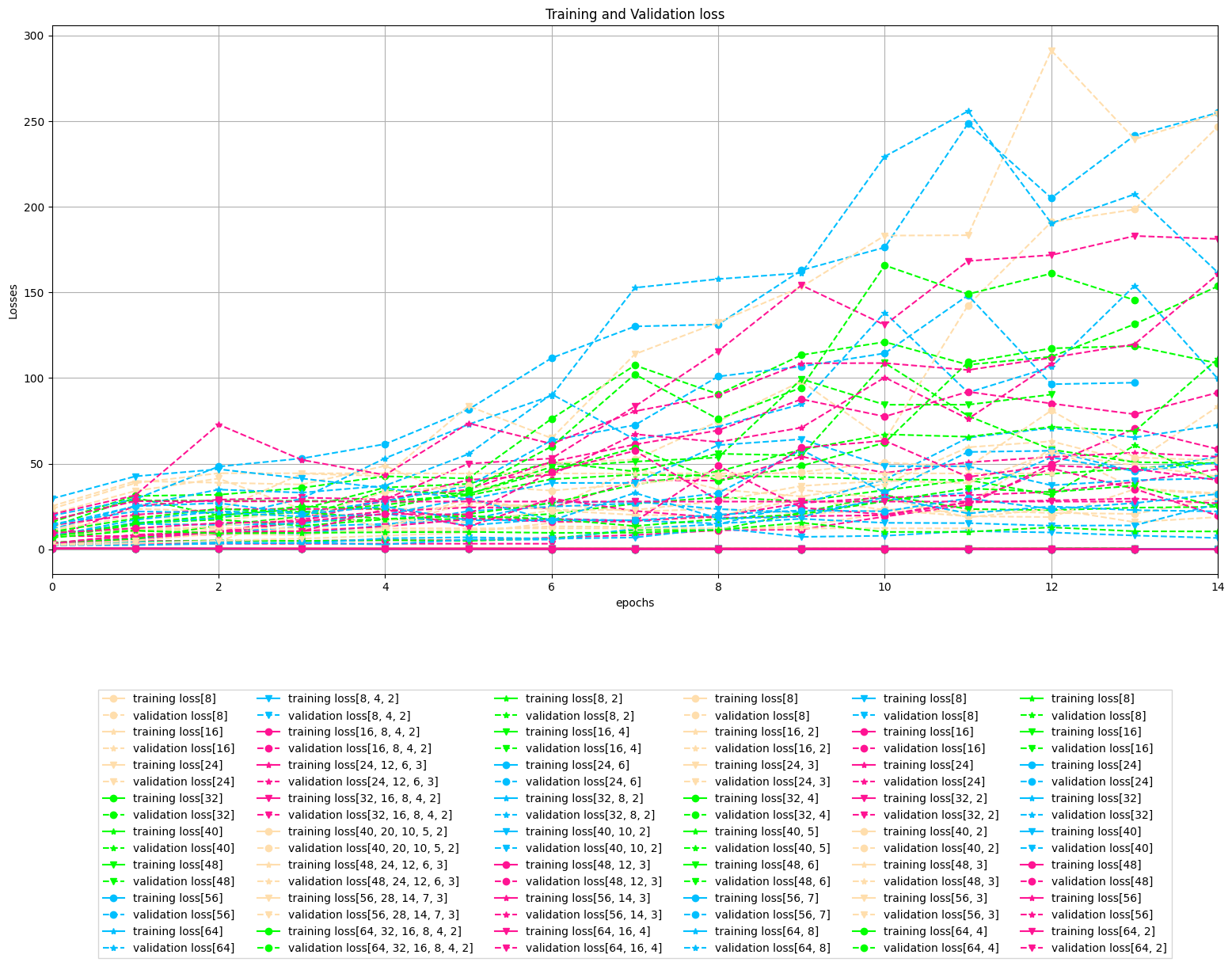}}
\caption{Loss Plots of Models found by Brute Force Traversal for Churn Model}
\label{churn_brute_force_search_loss}
\end{figure}

\begin{figure}[htbp!]
\centerline{\includegraphics[width=\linewidth]{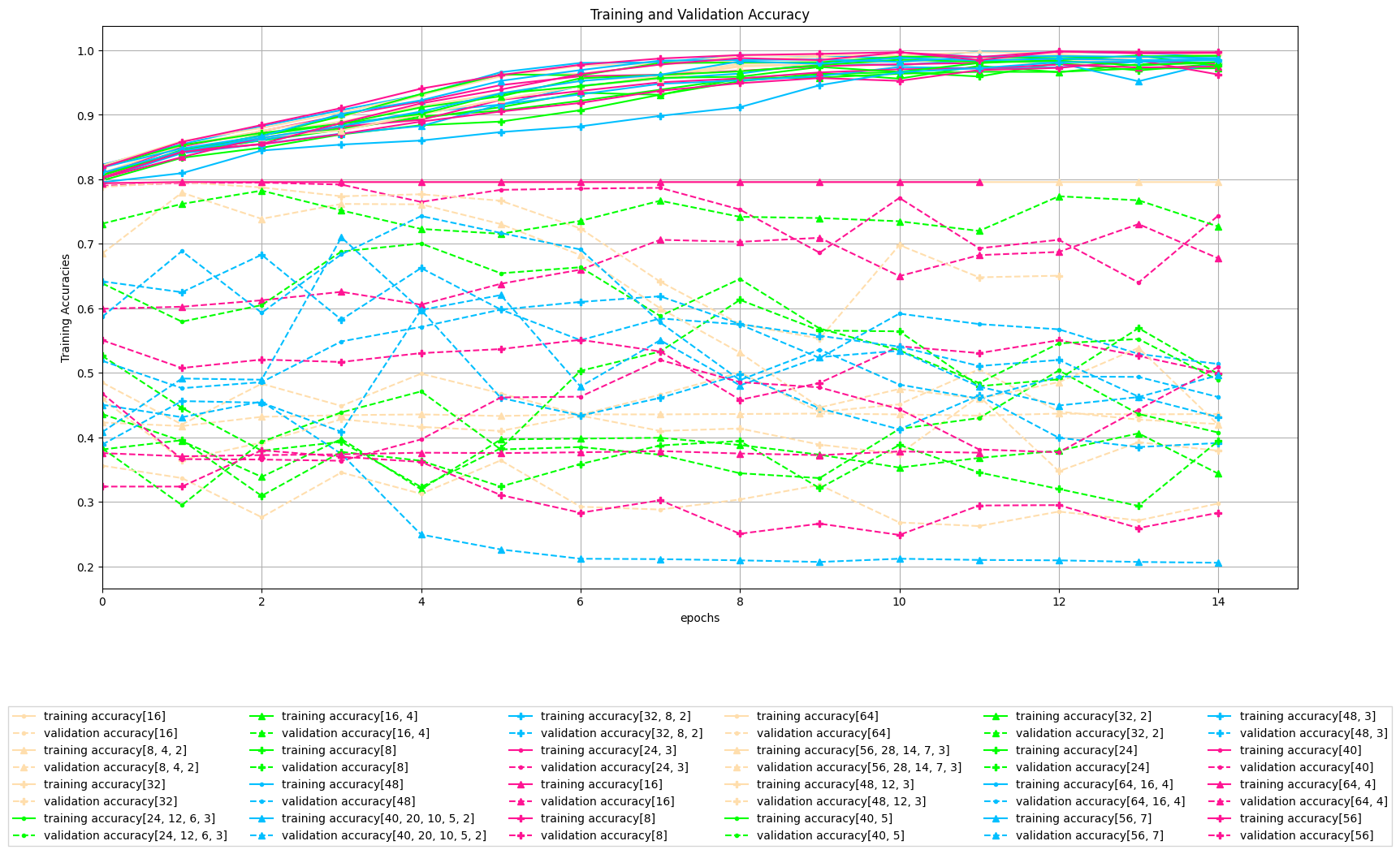}}
\caption{Accuracy Plots of Models found by Diagonal Traversal for Churn Model}
\label{churn_diagonal_search_accuracy}
\end{figure}

\begin{figure}[htbp!]
\centerline{\includegraphics[width=\linewidth]{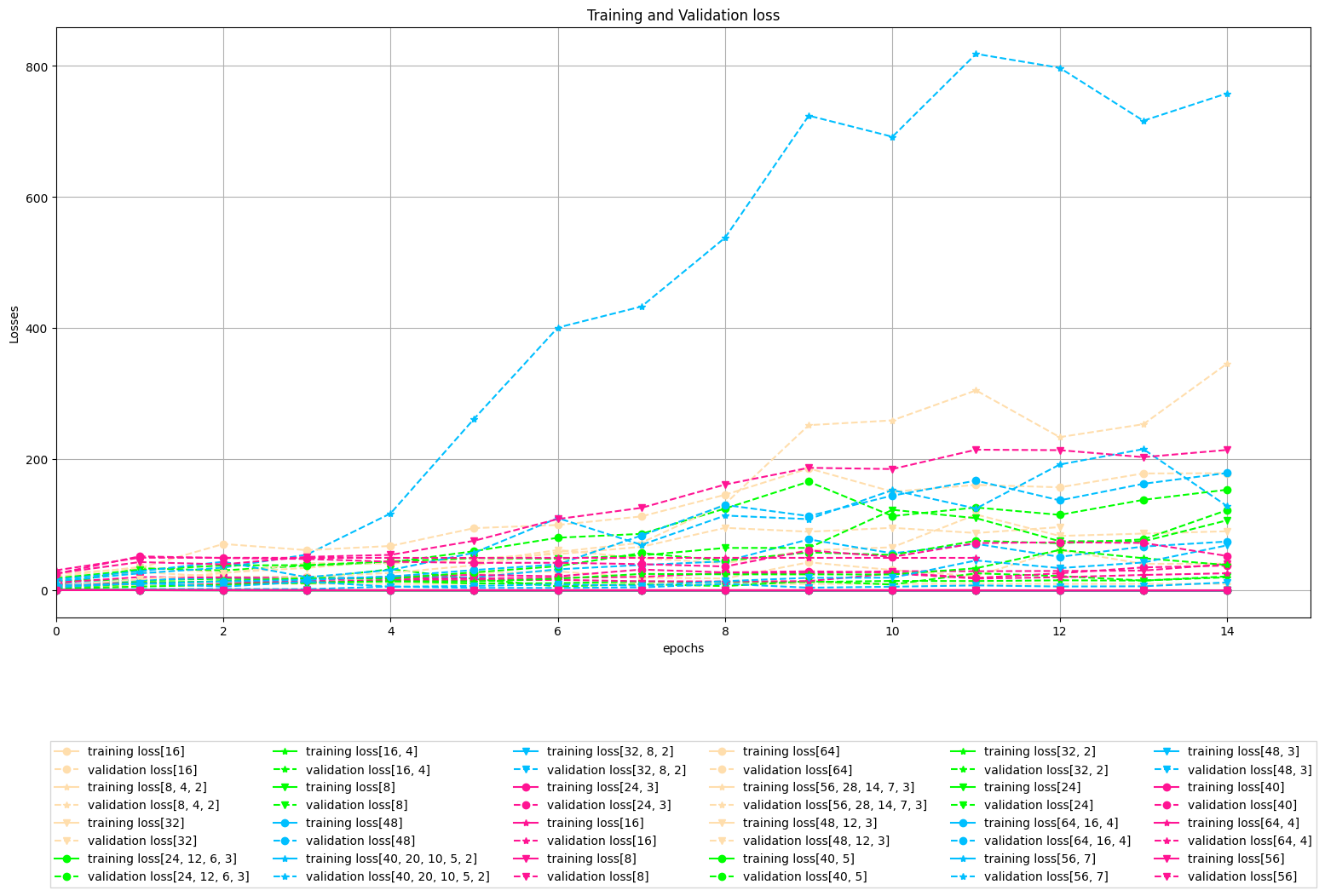}}
\caption{Loss Plots of Models found by Diagonal Traversal for Churn Model}
\label{churn_diagonal_search_loss}
\end{figure}

\end{document}